\documentclass[a4paper,conference]{IEEEtran}
\ifCLASSINFOpdf
\else
\fi
\usepackage{times}
\usepackage{graphicx}
\usepackage{amsmath}
\usepackage{amsthm}
\usepackage{booktabs}
\usepackage{algorithm}
\usepackage{algorithmic}
\usepackage{multirow}

\usepackage{amssymb}
\usepackage{caption,subcaption}

\begin{document}
%
\title{RGB-IR Cross-modality Person ReID based on Teacher-Student GAN Model}

\author{\IEEEauthorblockN{Ziyue Zhang,
Shuai Jiang,
Congzhentao Huang,
Yang Li and
Richard Yi Da Xu}
\IEEEauthorblockA{Faculty of Engineering and IT\\
University of Technology Sydney\\
81 Broadway, Sydney, Australia}}




%


\maketitle


\begin{abstract}
RGB-Infrared (RGB-IR) person re-identification (ReID) is a technology where the system can automatically identify the same person appearing at different parts of a video when light is unavailable. The critical challenge of this task is the cross-modality gap of features under different modalities. To solve this challenge, we proposed a Teacher-Student GAN model (TS-GAN) to adopt different domains and guide the ReID backbone to learn better ReID information. (1) In order to get corresponding RGB-IR image pairs, the RGB-IR Generative Adversarial Network (GAN) was used to generate IR images.  (2) To kick-start the training of identities, a ReID Teacher module was trained under IR modality person images, which is then used to guide its Student counterpart in training. (3) Likewise, to better adapt different domain features and enhance model ReID performance, three Teacher-Student loss functions were used. Unlike other GAN based models, the proposed model only needs the backbone module at the test stage, making it more efficient and resource-saving. To showcase our model’s capability, we did extensive experiments on the newly-released SYSU-MM01 RGB-IR Re-ID benchmark and achieved superior performance to the state-of-the-art with $49.8\%$ Rank-1 and $47.4\%$ mAP.
\end{abstract}


%
\IEEEpeerreviewmaketitle

\section{Introduction}

Person ReID is also referring to as pedestrian ReID. It is designed to match specific pedestrians in images or video sequences. Given a pedestrian image from one camera view, the algorithm tries to identify the same pedestrian from another view. The main challenge of ReID is that the intra-class (same person in different situations) variations are usually significant due to the changes in camera viewing conditions, such as viewpoint or situation differences, which makes it difficult to identify the same person. Meanwhile, the inter-class (different people in the same situation) variations also influence ReID performance. From a distance as typically in surveillance videos, people can look almost identical if they are in the same clothes and have similar body shapes. In summary, different pedestrians may appear to be similar, while the same pedestrian may look distinguishably at different times and locations.

In recent years, most existing works in person ReID are to learn discriminative features of person identity by a specifically designed backbone model \cite{chang2018multi,li2018harmonious,liu2017hydraplus,si2018dual,xu2018attention,yang2019towards,zhou2019osnet}. There are also works focusing on problems of occlusion factors in pedestrians' variations \cite{hou2019vrstc, huang2018adversarially}, different poses \cite{qian2018pose, zhu2019progressive} and views and resolutions \cite{li2019recover}.
These works are under the RGB-RGB camera setting. Both query images and the gallery images are in RGB mode.

\begin{figure}[!h]
    \centering
    \includegraphics[width = 0.45\textwidth]{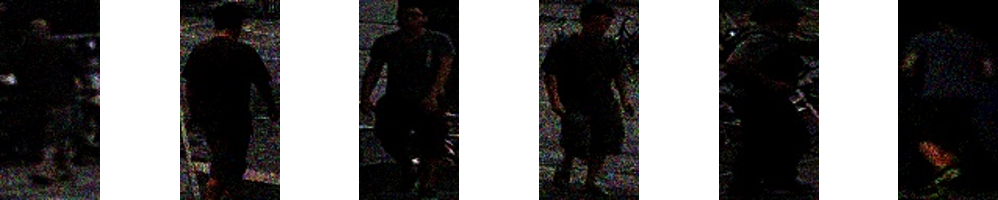}
    \caption{Example of RGB person images at night.}
    \label{fig:DarkExample}
\end{figure}

However, RGB-RGB camera ReID is greatly restricted when the light condition is weak or unavailable. A person may appear in one camera during the day and reemerge in another camera at night. In such a case, the RGB images captured at night by an RGB camera will have little effective information in ReID because of the darkness. As shown in Fig \ref{fig:DarkExample}, human eyes can hardly get any person identity information in the images which have lots of noise as well.

\begin{figure}[!h]
    \centering
    \includegraphics[width = 0.45\textwidth]{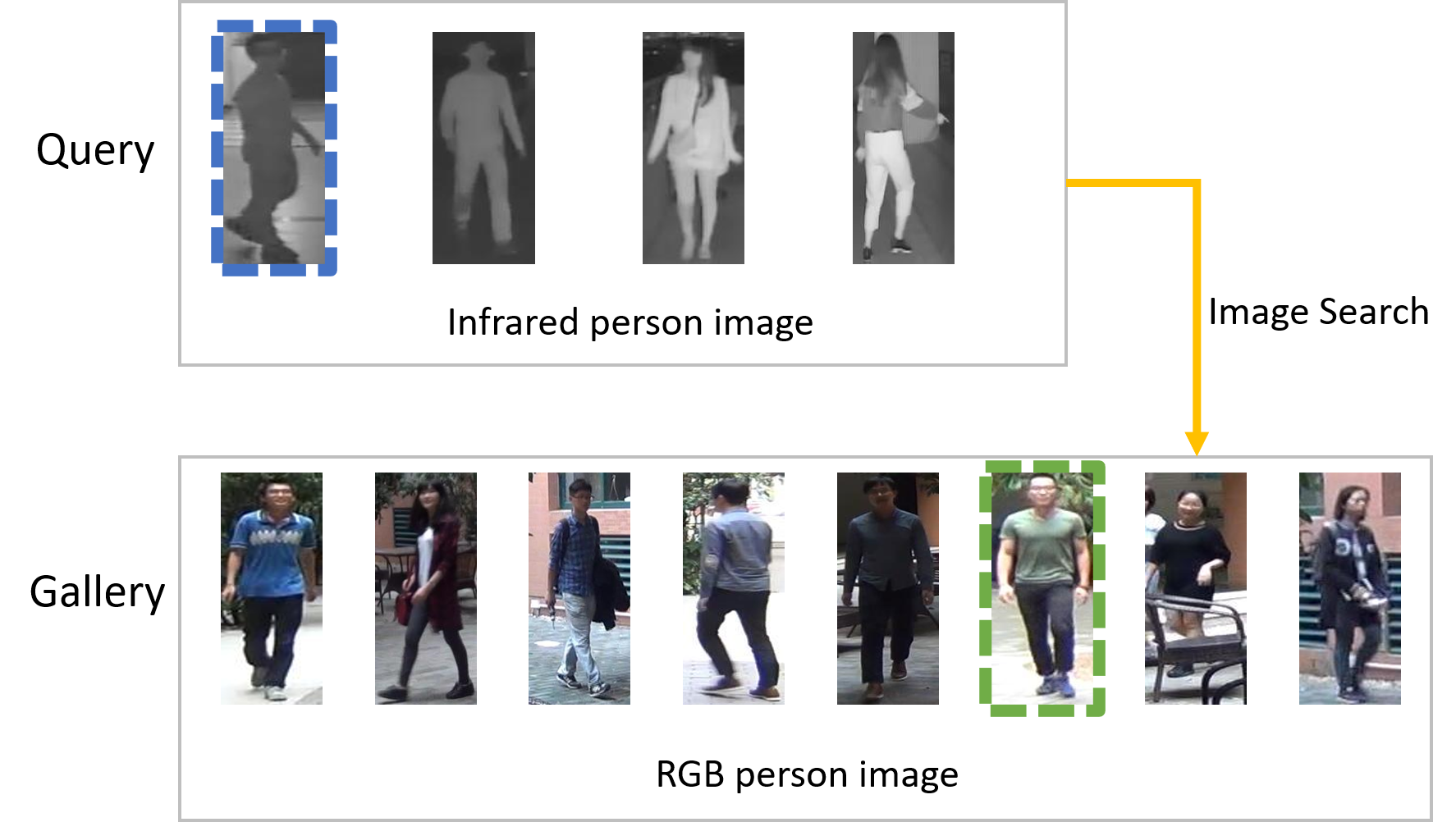}
    \caption{The example of RGB-IR cross-modality ReID. The query images are under IR modality and the gallery images are under RGB modality. The person with blue rectangle in the query and the person with green rectangle in the gallery have the same identity.}
    \label{fig:ReidExample}
\end{figure}

As known to all, an infrared camera forms a grey image (single channel image) using infrared radiation, which can increase in-the-dark visibility without actually using a visible light source. Thus, using both RGB and IR images will complement each other and enhance person ReID performance. As shown in Fig \ref{fig:ReidExample}, the query images are all under IR modality, providing much more information than those in Fig \ref{fig:DarkExample}.

However, few researchers have studied such RGB-IR cross-modality person ReID. The main challenge is that different pedestrians can appear to be very similar in the same modality, while the same pedestrian under different modalities can look quite different. Another challenge is that IR images only have grey-scale pixels, which provide much less information comparing to RGB images, making it more challenging to extract effective features for the task of ReID. 

To resolve the above two challenges, some studies simply use the share-weights ReID backbone for both RGB and IR domain inputs to get identity features in the same latent space \cite{wu2017rgb}. Some use independent ReID backbones for each domain input to constrain identity features by the share-weights classifier and triplet loss \cite{ye2018hierarchical}. These works with no use of GAN cannot well solve the domain gap only by the guidance of shared or independent ReID backbone due to the lack of pair information. To better minimize the domain gap, some other works use GAN to generate synthetic IR images of people using corresponding RGB inputs \cite{dai2018cross,wang2019rgb}. They require the usage of GAN at both the train and test stage, which is resource-wasting and time-consuming.

Different from these works, in this paper, we propose a Teacher-Student GAN based cross-modality person ReID model (TS-GAN). The critical insight of our approach is that we design a novel network in which the Student ReID module is guided by a pretrained Teacher module to encourage the closeness between RGB and IR ReID features. It tremendously improves the quality of features obtained for ReID classification and reduces the gap between different modalities. To improve the model's effectiveness, we have added several innovations summarised as follows:

\begin{enumerate}
\item
IR ReID Teacher module is pretrained by using Real IR images in the train-set, which obtains very high accuracy. We then use it as the teacher to guide the feature learning in the Student ReID module.
\item 
We use cycle-consistency GAN with joint discriminator to generate the corresponding Fake IR person images from the input Real RGB person images, thus obtain pair-wise person images under different modalities. The (Real RGB, Fake IR) image pairs are then used to train the Student ReID module with MSE reconstruction loss such that the cross-modality gap can be reduced.
\item 
To enhance the feature extraction ability of the Student module, we also use another two MSE losses. One is between Real IR image features from the Teacher module and those from the Student module. The other one is between Fake IR image features from both modules.
\item
Unlike other GAN based methods in cross-modality tasks, our model only requires the GAN module at the train stage. During testing, it is more resource-saving and efficient to feed-forward through the ReID backbone module without the involvement of GAN.
\end{enumerate}

In order to demonstrate the effectiveness of our algorithm, extensive experiments were carried out on an RGB-IR Cross-modality ReID datasets, SYSU-MM01.


\section{Related work}

\subsection{RGB-RGB Person ReID}

Most researchers focused on traditional RGB-RGB person ReID. One primary method of person ReID is metric learning, which is to formalize the problem as supervised metric learning where a projection matrix is sought out \cite{yang2017person}. 
By using a Siamese network, Yi et al. \cite{yi2014deep} proposed a method that can jointly learn the color feature, texture feature and metric in a unified framework. 
Ding et al. \cite{ding2015deep} proposed a deep image representation based on a scalable distance comparison for person Re-ID. 
Another primary method is to learn appropriate features associated with the same ID using features distance information \cite{hermans2017defense} on a backbone module, such as Resnet50 \cite{he2016deep}. 
Yang et al. \cite{yang2019towards} proposed a Class Activation Maps augmentation model to expand the activation scope of the baseline ReID model to explore rich visual cues.
Chang et al. \cite{chang2018multi} proposed Multi-Level Factorisation Net, which factorizes the visual appearance of a person into latent discriminative factors at multiple semantic levels without manual annotation.
Zhao et al. \cite{zhao2017spindle} proposed Spindle Net to extract multiple semantic region features and merge them to get whole person discriminative features.
All of these works focus on RGB-RGB person ReID, which may fail in some circumstances, e.g., when RGB images cannot provide sufficient features at night. They cannot handle multi-modality data, which will lead to a heterogeneous gap.

\subsection{RGB-IR Cross-modality Person ReID}
In addition to the difficulty of traditional person ReID, cross-modality person ReID faces the problem that the characteristics of the same person in different modalities can be very different. To address this issue, researchers mainly focused on how to reduce the gap between different modalities.
Wu et al. \cite{wu2017rgb} proposed a deep zero-padding network for cross-modality matching and contributed a new multiple modalities ReID dataset SYSU-MM01. 
Ye et al. \cite{ye2018hierarchical} proposed a two-stream network with feature learning and metric learning for transforming two heterogeneous modalities into a consistent space that modality-shared metric.
Feng et al. \cite{feng2019learning} proposed a framework that employed modality-specific networks to tackle the heterogeneous matching problem.
Dai et al. \cite{dai2018cross} proposed a cross-modality generative adversarial network (termed cmGAN) to handle the lack of insufficient discriminative information.
Wang et al. \cite{wang2019learning} introduced a Dual-level Discrepancy Reduction Learning scheme which handles the two discrepancies separately.
Wang et al. \cite{wang2019rgb} proposed an end-to-end Alignment Generative Adversarial Network (AlignGAN) for the RGB-IR ReID task. 
In our work, we gather the middle layer feature map information to adapt the difference between RGB and IR modality, which is unique in this area.

\subsection{GAN in Person ReID}
GAN provides a way to learn deep representations without considerable annotated data. The representation learned by GAN can be used in a variety of applications, including person ReID.
The first GAN was proposed in 2014 \cite{NIPS2014_5423}. After that, researchers developed multiple task-specific GAN structures, such as Pix2Pix \cite{isola2017image} and cycleGAN \cite{zhu2017unpaired} in style translation. 
There are also many works in the field of person ReID using GAN to improve its effectiveness.
Some GAN based methods \cite{zheng2017unlabeled,zheng2019joint} were proposed to augment the training data and enhance the tolerance to deal with the invariance of input changes. Li et al. \cite{li2019recover} proposed a network allowing query images with different resolutions to address cross-resolution person ReID.
However, these methods cannot be adopted directly on the RGB-IR cross-modality domain adaptation problem. Hence, we only use the idea of cycle consistency GAN to support our model's training process.

\subsection{Teacher-Student Model in Cross-modality Tasks}
Teacher-Student (TS) model has been widely used in compressing deep learning models \cite{hinton2015distilling,chen2018darkrank}, which is a widespread practice in the field of model compression. It first learns the Teacher network by building a relatively complex network structure using the training dataset and then re-predicts the training data set using the learned Teacher network to guide the simpler Student network. 
There are also some works using the TS model in general cross-modality tasks. Zhao et al. \cite{zhao2018through} used RGB Teacher to guide the wireless Student in the wireless networking field. Thoker et al. \cite{thoker2019cross} used the TS model to transfer a modality that can be adapted to recognize actions for another modality, such as from RGB videos to sequences of 3D human poses. 
We use the TS model in our method to reduce the domain gap between different modalities. By setting one modality model as the Teacher module, we can guide the cross-modality Student module's feature learning ability.

\section{Method}

\begin{figure*}[!ht]
    \centering
    \includegraphics[width = 1\textwidth]{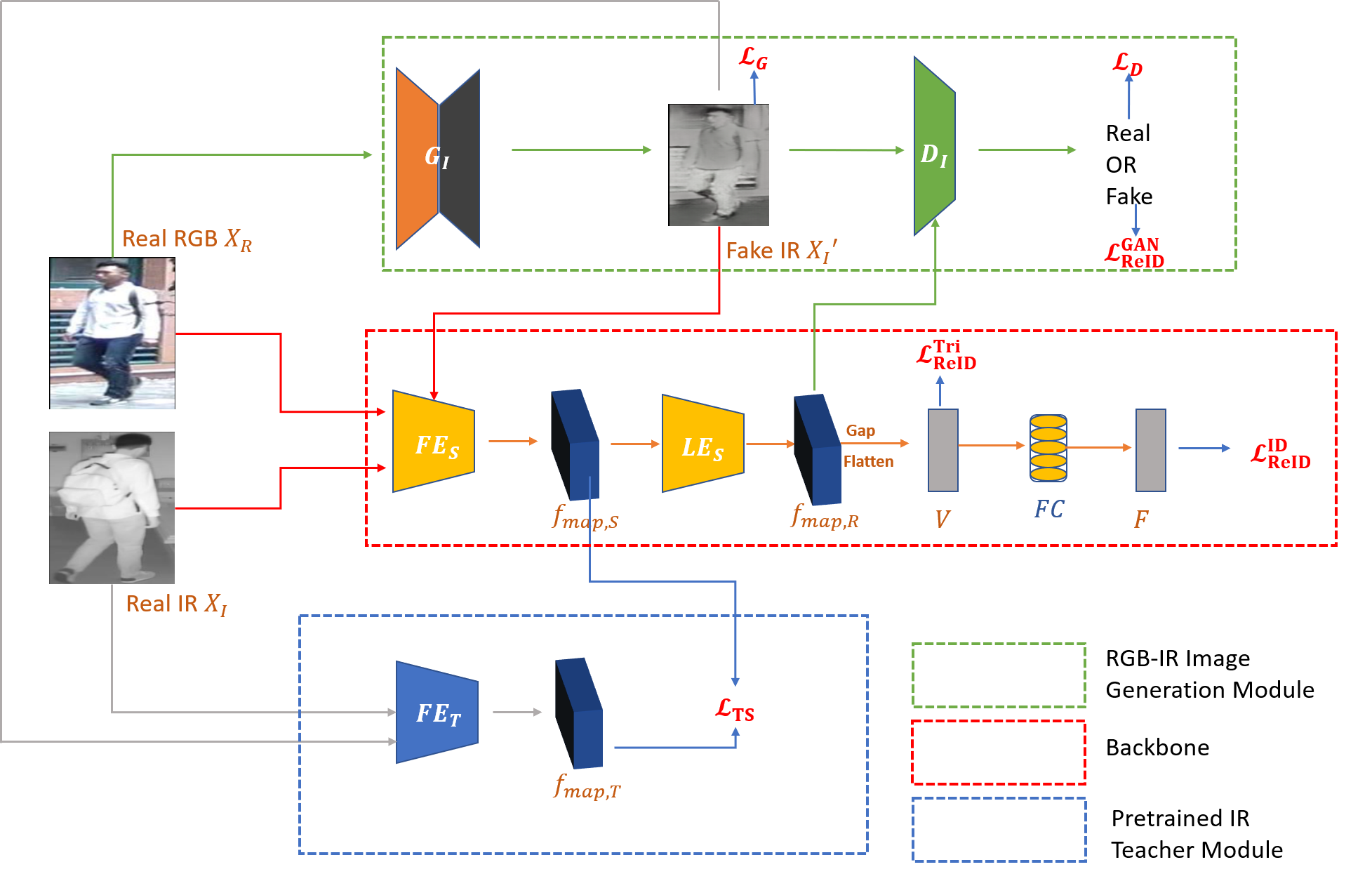}
    \caption{ The whole model structure of our cross-modality person ReID task (best viewed in color). $G_I$ and $D_I$ (in the green dotted rectangle) are the generator and joint discriminator for IR images. $FE_S$ and $LE_S$ (in the red dotted rectangle) are the former encoder and the latter encoder for the Student ReID backbone. ${FE}_T$ (in the blue dotted rectangle) is the pretrained IR ReID Teacher module. The inputs of the ReID backbone are Real RGB and Real IR person images as well as the generated Fake IR person images. The inputs of the IR ReID Teacher module are Real IR and Fake IR person images.}
    \label{fig:wholeModels}
\end{figure*}

The overall model structure for our TS-GAN model is shown in Fig \ref{fig:wholeModels}.
The whole model consists of three main parts, which are: (1) RGB-IR image generation module, (2) ReID backbone and (3) RGB-IR TS module. We use subscripts ``S'' and ``T'' to distinguish blocks belonging to the Student or Teacher modules.

At the train stage, these three components will work together to train the cross-modality person ReID backbone module. The inputs are images under different modalities and the model outcomes are a person ReID feature and a generated person image under another modality. The main output during testing is the person ReID feature.

\subsection{RGB-IR Image Generation Module}

To address the cross-modality variation, we construct a cycle consistency GAN to generate Fake IR images from corresponding Real RGB images. 
The module consists of one generator $G_{I}$ for transferring RGB images to IR images and its corresponding joint discriminator $D_{I}$. 
The input of the generator $G_{I}$ is no longer the random noise as ordinary GAN, but the Real RGB images $X_{R}$. The goal of the generator $G_{I}$ is to make the generated images $X_{I}'$ and the Real IR images $X_{I}$ as similar as possible. Different from ordinary discriminators, the input of our joint discriminator $D_{I}$ is a pair of IR image and ReID feature map. The adversarial loss of IR generation is defined as:
\begin{equation}
\label{l-gan-ir-gen}
\mathcal{L}_{G}^{\text{adv}} = \mathbb{E}\left[\log D_{I}(X_{I}', f_{\text{map}}^{X_{I}'}))\right],
\end{equation}
\begin{equation}
\label{l-gan-whole-dis}
\mathcal{L}_{D} = \mathcal{L}^{\text{real}}_{D} + \mathcal{L}^{\text{fake}}_{D},
\end{equation}
where
\begin{equation}
\label{l-gan-ir-real-dis}
\mathcal{L}^{\text{real}}_{D} = \mathbb{E}_{(x,f) \in (X_{I},f_{\text{map,R}}^{X_I})}\left[\log D_{I}(x,f)\right],
\end{equation}
\begin{equation}
\label{l-gan-ir-fake-dis}
\mathcal{L}^{\text{fake}}_{D} = \mathbb{E}_{(x,f) \in M}\left[\log \left(1-D_{I}(x,f)\right)\right],
\end{equation}
\begin{equation}
\label{M-defination}
M = (X_{I}', f_{\text{map,R}}^{X_{I}'}) \cup (X_{I}', f_{\text{map,R}}^{X_{I}}) \cup (X_{I}, f_{\text{map,R}}^{X_{I}'}),
\end{equation}
$f_{\text{map,R}}^{X_{I}}$ and $f_{\text{map,R}}^{X_{I}'}$ are the ReID feature maps of input $X_{I}$ and generated $X_{I}'$ respectively.

Eq. (\ref{l-gan-ir-gen}) is for training the generator while Eq. (\ref{l-gan-ir-real-dis}) and (\ref{l-gan-ir-fake-dis}) are for training the discriminator. Using the joint discriminator rather than the ordinary discriminator, we can further enhance the ability of the backbone module and discriminator module to reduce the cross-modality gap. However, these losses are insufficient because the generator $G_{I}$ can completely map all Real RGB images $X_{R}$ to the same picture in IR modality and make the loses invalid. Hence to further support the training process of IR image generation, we also introduce the generator $G_{R}$ for transferring IR images to RGB images and its corresponding discriminator $D_{R}$. For conciseness, we do not show their loss definitions here. Therefore, the cycle-consistency loss we used is:


\begin{equation}
\label{l-gan-cyc}
\begin{aligned}
\mathcal{L}_{\text{cyc}}& = E[\|G_{R}(G_{I}(X_{R}))-X_{R}\|_{1}] \\
& + E[\|G_{I}(G_{R}(X_{I}))-X_{I}\|_{1}],
\end{aligned}
\end{equation}
where we use the L1 norm as it can achieve better generation performance at the edge of images than the L2 norm. This loss is to make Fake IR images generated by $G_{I}$ consistent with the input Real RGB image in content. Specificly, after obtaining the Fake IR by inputing a Real RGB, we send the Fake IR to the RGB generator and obtain a reconstructed RGB. By constraining the reconstructed RGB and input RGB to be the same, we can make sure that the content in both the Fake IR and Real RGB images is consistent.

Now the updated generator loss function can be written as follows:
\begin{equation}
\label{l-gan-whole-gen}
\mathcal{L}_{G} = \mathcal{L}_{G}^{\text{adv}} + \omega * \mathcal{L}_{\text{cyc}},
\end{equation}
where $\omega$ is the weight of cycle loss. By using this loss in adversarial training process, we can generate IR images in high quality for latter ReID and TS module.

\subsection{ReID Backbone Module}

The whole backbone is based on the baseline model \cite{Luo_2019_CVPR_Workshops}. It takes RGB-IR cross-modality images ($X_{R}$ and $X_{I}$) and the generated IR images $X_{I}'$ as inputs such that the backbone can perform better ReID when querying Real RGB images. The backbone consists of two parts. The first part is the former encoder ($\text{FE}_{S}$) for shallow feature extraction, which is the first convolution layer followed by the first three blocks of Resnet50. The former encoder will output the ReID middle layer feature maps $f_{\text{map,S}}$. The second part is the latter encoders ($\text{LE}_{S}$) for extracting person ReID features, which is the last block of Resnet50. Then we adopt a Global Average Pooling (GAP) layer on the ReID feature map (outputs of the latter encoders) and then flatten them to get the person ReID latent vector ($V$). Finally, we use a fully connected layer (FC) to convert $V$ to the person ReID factor ($F$).
 
To classify the person's identity, we adopt three loss functions. The first one is the soft-label identity classification loss $\mathcal{L}_{\text{ReID}}^{\text{ID}}$ for $F$. The second one is the triplet loss $\mathcal{L}_{\text{ReID}}^{\text{Tri}}$ for $V$ and the last one is the GAN loss $\mathcal{L}_{\text{ReID}}^{\text{GAN}}$. Hence the complete person ReID loss function is defined as:
\begin{equation}
\label{l-reid-whole}
\mathcal{L}_{\text{ReID}} =\lambda_{1} \mathcal{L}_{\text{ReID}}^{\text{ID}}+\lambda_{2} \mathcal{L}_{\text{ReID}}^{\text{Tri}} + \lambda_{3} \mathcal{L}_{\text{ReID}}^{\text{GAN}},
\end{equation}
where $\lambda_{1}$, $\lambda_{3}$ and $\lambda_{3}$ are the weights allocated for the above three loss functions. 

$\mathcal{L}_{\text{ReID}}^{\text{ID}}$ is the classification loss using soft label for person ReID. For the $i$th data in a batch, it is defined as:
\begin{equation}
\label{l-reid-cls}
\mathcal{L}_{\text{ReID}}^{\text{ID}} = -\sum_{n=1}^{N} q_{i,n} \log(p_{i,n}),
\end{equation}
where $N$ is the number of person identities (logits) and $p_{i,n}$ is the prediction logit for the $n$th identity.
As for $q_{i,n}$, since person ReID is a few-shot learning problem, we adopt the soft label version to prevent over-fitting, i.e.
\begin{equation}
\label{l-reid-cls-sup}
    q_{i,n}=
    \begin{cases}
    	1-\frac{N-1}{N} \epsilon & \text{if $i=n$} \\ \epsilon / N & \text{otherwise}
    \end{cases},
\end{equation}
where $\epsilon$ is a tiny weight factor allocated on those false label classes in the training dataset.

$\mathcal{L}_{\text{ReID}}^{\text{Tri}}$ is the batch hard triplet loss for person ReID vector ($V$) similarity learning. It makes the distance of vectors closer between the same identity and further apart between different identities. The batch hard triplet loss function can be defined as follows:
\begin{equation}
\label{l-reid-tri}
\mathcal{L}_{\text{ReID}}^{\text{Tri}} = \sum_{\alpha, p, n \atop y_{a}=y_{p} \neq y_{n}}\left[m+D_{a, p}-D_{a, n}\right]_{+},
\end{equation}
where $y_{a}, y_{p}, y_{n}$ is the corresponding truth ID of sample $a,p,n$. $D_{a, p}$ and $D_{a, n}$ are the Euclidean distance between the anchor vector $V_a$ and positive sample vector $V_p$ (same identity sample) and that between the anchor vector $V_a$ and negative sample vector $V_n$ (different identity sample), $m$ is a margin parameter constraining the maximal distance between the anchor and negative samples and $[x]_{+} = \max(0, x)$.

$\mathcal{L}_{\text{ReID}}^{\text{GAN}}$ is used to fool the joint discriminator module and defined as
\begin{equation}
\label{l-reid-gan}
\mathcal{L}_{\text{ReID}}^{\text{GAN}} = \mathbb{E}_{(x,f) \in M}\left[\log D_{I}(x,f)\right].
\end{equation}
where $M$ is defined in Eq. (\ref{M-defination}).

\subsection{RGB-IR Teacher-Student Module}

Since image features associated with different modalities are complicated for the encoder to disentangle, we construct an IR ReID Teacher module to guide this process on the middle layer features between $\text{FE}_{S}$ and $\text{LE}_{S}$. Here we use IR person images as inputs of the Teacher module rather than RGB person images because the quality of generated IR images is much higher than that of generated RGB images. Although IR images lack color information, the Teacher is still able to get very high accuracy (R1 98\%, see Experiments) when both query and gallery images are under IR modality. That means IR images have sufficient information on person identities for classification.
The ReID Teacher module consists of only the former encoder ($\text{FE}_{T}$), which is to extract features from Real IR images and Fake IR images during training. We pretrain a baseline model using Resnet50 on the IR images of SYSU-MM01 training set by the triplet loss and identification classification loss. After pretraining, we fix its weights. The ReID Teacher module is comprised of the first three blocks of the pretrained fix-weights Resnet50 baseline model.

During training, the ReID Teacher module takes the Real IR images $X_{I}$ and generated Fake IR images $X_{I}'$ as inputs. 
To make the guidance from Teacher module to Student ReID backbone module, we construct a regularization loss between the output of the ReID Teacher module ($f_{\text{map,T}}^{X_{I}}$ and $f_{\text{map,T}}^{X_{I}'}$) and the output of the ReID former encoder ($f_{\text{map,S}}^{X_{I}}$, $f_{\text{map,S}}^{X_{R}}$ and $f_{\text{map,S}}^{X_{I}'}$). We use two MSE losses to minimize the difference between these two feature maps such that the gap between the Student backbone module and IR Teacher module can be reduced:
\begin{equation}
\label{l-TS-realir}
\mathcal{L}_{\text{TS}}^{\text{RealIR}}=\lVert f_{\text{map,T}}^{X_{I}} - f_{\text{map,S}}^{X_{I}}\rVert_{2}^{2},
\end{equation}
\begin{equation}
\label{l-TS-fakeir}
\mathcal{L}_{\text{TS}}^{\text{FakeIR}}=\lVert f_{\text{map,T}}^{X_{I}'} - f_{\text{map,S}}^{X_{I}'}\rVert_{2}^{2}.
\end{equation}
To adapt different modality features, we use the following cross-domain loss
\begin{equation}
\label{l-TS-cross}
\mathcal{L}_{\text{TS}}^{\text{CD}}=\lVert f_{\text{map,T}}^{X_{I}'} - f_{\text{map,S}}^{X_{R}}\rVert_{2}^{2},
\end{equation}
which can reduce the domain gap between RGB and IR modality by encouraging the middle layer feature maps of RGB input being close to those of IR inputs. Because of this loss function, the Student ReID backbone module can learn IR person information, such as textures and gestures, from RGB person images.

The whole TS ReID loss is as follows:
\begin{equation}
\label{l-TS-whole}
\mathcal{L}_{\text{TS}}=\alpha_{c} * \mathcal{L}_{\text{TS}}^{\text{CD}} + \alpha_{s} * (\mathcal{L}_{\text{TS}}^{\text{RealIR}} + \mathcal{L}_{\text{TS}}^{\text{FakeIR}}),
\end{equation}
where $\alpha_c$ and $\alpha_s$ are weights for the cross-domain TS loss and the same-domain TS loss.

\subsection{Training and Testing}

In our setting, we choose $P$ RGB-IR person image pairs for each of $K$ identities in trainset randomly as a mini-batch. So the batch size is $2*P*K$. Then we send our mini-batch data to our TS-GAN model to train all components alternatively. Here we define the optimizer of joint discriminator, IR generator and ReID backbone as $O_{D}$, $O_{G}$ and $O_{B}$. 
Below we give the pseudo-code about how the updates are performed on TS-GAN model in one step.

\begin{algorithm}[h]
\caption{Training process of TS-GAN for $1$ step}
\label{alg:algorithm}

\begin{algorithmic}[1] 
\REQUIRE  $P$ real RGB images $X_{R}$ and $P$ real IR images $X_{I}$\\
\ENSURE $\omega, \lambda_{1}, \lambda_{2}, \lambda_{3}, \alpha_{c}, \alpha_{s}$ \\
\STATE Feed forward $X_{R}$ to IR generator and get fake images $X_{I}'$. \\ Calculate $\mathcal{L}_G$ as Equation \ref{l-gan-whole-gen} and backward.
\STATE Update generator parameters $O_{G}$ with $\mathcal{L}_G$.
\STATE Feed forward $X_{R}, X_{I}, X_{I}'$ to Backbone backbone module and fixed weights IR Teacher module.\\Calculate the whole loss for ReID backbone $\mathcal{L}_{\text{Backbone}} = \mathcal{L}_{\text{ReID}} + \mathcal{L}_{\text{TS}}$ and backwards.
\STATE Update ReID backbone parameters $O_{B}$ with $\mathcal{L}_{\text{Backbone}}$.
\STATE Feed forward $X_{R}, X_{I}, X_{I}'$ and corresponding feature maps to joint discriminator.
\STATE Calculate $\mathcal{L}_D$ as Equation \ref{l-gan-whole-dis} and backward.
\STATE Update discriminator parameters $O_{D}$ with $\mathcal{L}_D$.
\end{algorithmic}
\end{algorithm}

As shown in Algorithm \ref{alg:algorithm}, the training process for each step consists of three phases. 
They are trained alternatively and all components of TS-GAN model are used at the train stage.
However, at the test stage, only the second phase is needed for getting person ReID vector $V$, which makes our model efficient and resource-saving.

\begin{table*}[!ht]

	\centering
	\caption{Comparison with state-of-the-art ReID models on SYSU-MM01.The R1, R10 and R20 are Rank-1, Rank-10 and Rank-20 accuracies (\%). The mAP denotes mean Average Precision score (\%).}
	\resizebox{\textwidth}{!}{%
		\begin{tabular}{lcccccccccccccccc}
			\hline
			\multirow{3}{*}{Method} & \multicolumn{8}{c}{All-Search} & \multicolumn{8}{c}{Indoor-Search} \\ \cline{2-17} 
			& \multicolumn{4}{c}{Single-Shot} & \multicolumn{4}{c}{Multi-Shot} & \multicolumn{4}{c}{Single-Shot} & \multicolumn{4}{c}{Multi-Shot} \\ \cline{2-17} 
			& R1 & R10 & R20 & mAP & R1 & R10 & R20 & mAP & R1 & R10 & R20 & mAP & R1 & R10 & R20 & mAP \\ \hline
			HOG & 2.76 & 18.3 & 32.0 & 4.24 & 3.82 & 22.8 & 37.7 & 2.16 & 3.22 & 24.7 & 44.6 & 7.25 & 4.75 & 29.1 & 49.4 & 3.51 \\
			LOMO & 3.64 & 23.2 & 37.3 & 4.53 & 4.70 & 28.3 & 43.1 & 2.28 & 5.75 & 34.4 & 54.9 & 10.2 & 7.36 & 40.4 & 60.4 & 5.64 \\
			Two-Stream\cite{wu2017rgb} & 11.7 & 48.0 & 65.5 & 12.9 & 16.4 & 58.4 & 74.5 & 8.03 & 15.6 & 61.2 & 81.1 & 21.5 & 22.5 & 72.3 & 88.7 & 14.0 \\
			One-Stream\cite{wu2017rgb} & 12.1 & 49.7 & 66.8 & 13.7 & 16.3 & 58.2 & 75.1 & 8.59 & 17.0 & 63.6 & 82.1 & 23.0 & 22.7 & 71.8 & 87.9 & 15.1 \\ 
			Zero-Padding\cite{wu2017rgb} & 14.8 & 52.2 & 71.4 & 16.0 & 19.2 & 61.4 & 78.5 & 10.9 & 20.6 & 68.4 & 85.8 & 27.0 & 24.5 & 75.9 & 91.4 & 18.7 \\
			cmGAN\cite{dai2018cross} & 27.0 & 67.5 & 80.6 & 27.8 & 31.5 & 72.7 & 85.0 & 22.3 & 31.7 & 77.2 & 89.2 & 42.2 & 37.0 & 80.9 & 92.3 & 32.8 \\ 
			Dist.based\cite{Tekeli_2019_ICCV} & 29.05 & 74.71 & 87.16 & 30.94 & 35.40 & 81.02 & 91.85 & 24.12 & 32.74 & 82.40 & 93.35 & 44.26 & 40.41 & 86.83 & 96.27 & 33.93 \\ 
			alignGAN\cite{wang2019rgb} & 42.4 & 85.0 & 93.7 & 40.7 & 51.5 & 89.4 & 95.7 & 33.9 & 45.9 & 87.6 & 94.4 & 54.3 & 57.1 & \textbf{92.7} & 97.4 & 45.3 \\ 
			TS-GAN (w/o rerank) & \textbf{49.8} & \textbf{87.3} & \textbf{93.8} & \textbf{47.4} & \textbf{56.1} & \textbf{90.2} & \textbf{96.3} & \textbf{38.5} & \textbf{50.4} & \textbf{90.8} & \textbf{96.8} & \textbf{63.1} & \textbf{59.3} & 91.2 & \textbf{97.8} & \textbf{50.2} \\
			TS-GAN (w/ rerank) & \textbf{58.3} & \textbf{87.8} & \textbf{94.1} & \textbf{55.1} & \textbf{55.9} & \textbf{91.2} & \textbf{96.6} & \textbf{39.7} & \textbf{62.1} & \textbf{90.8} & \textbf{96.4} & \textbf{71.3} & \textbf{59.7} & 91.8 & \textbf{97.9} & \textbf{50.9} \\ \hline
		\end{tabular}%
	}
	
	\label{tab:table1}
\end{table*}
\section{Experiment}

\subsection{Dataset and Evaluation Protocol}
\subsubsection{SYSU-MM01 dataset}
SYSU-MM01 is the most popular and newest dataset in RGB-IR cross-modality person ReID. It contains images captured by six cameras, including two IR cameras and four RGB ones. Different from RGB cameras, IR cameras work in dark scenarios.
RGB images of camera 1 and camera 2 were captured in two bright indoor rooms (room 1 and room 2) by Kinect V1. For each person, there are at least 400 continuous RGB frames with different poses and viewpoints. IR images of camera 3 and camera 6 are captured by IR cameras in the dark. Camera 3 is placed in room 2 in a dark environment, while camera 6 is placed in an outdoor passage with background clutters. Camera 4 and 5 are RGB surveillance cameras placed in two outdoor scenes named gate and garden.

\subsubsection{Evaluation protocol}
We use the standard Cumulative Matching Characteristics (CMC) rank1 (R1), rank10 (R10), rank20 (R20) values and mean Average Precision (mAP) as the evaluation standards. Following \cite{wu2017rgb}, the results of SYSU-MM01 are evaluated by re-implemented code using our baseline framework.
There are two modes in the dataset. One is the all-search mode, where RGB cameras 1, 2, 4 and 5 are for gallery set and IR cameras 3 and 6 are for probe set. Another is the indoor-search mode, where RGB cameras 1 and 2 are for gallery set and IR cameras 3 and 6 are for probe set. We randomly choose one RGB image for the single-shot setting and ten RGB images for the multi-shot setting from the gallery set.

\subsection{Implementation Details}
We use one TESLA V100-SXM2-32GB graphics card for our entire experiments. The batch size is set as 64, including 4 RGB-IR image pairs per person from 8 people. We use Adam optimizer and set both the weight decay factor and weight decay bias factor as 0.0005. The base learning rate is 0.00035, with a linear learning rate scheduler. The total training epoch number is 120. For the epoch step in [40, 70], we decrease the learning rate by a decay factor 0.3.

For some hidden parameters in the train stage, we set weight parameters $\lambda_{1}$ as 3.0, $\lambda_{2}$ as 1.0 and $\lambda_{3}$ as 0.1. The triplet loss plays a more critical role in the person ReID results than identification loss since the number of people in the dataset is relatively small. We set the margin value $m$ in the triplet loss as 0.3. For the IR ReID TS model regularization loss, we set the weight factor $\alpha_s$ as 0.003 and $\alpha_c$ as 0.006. For the cycle-consistency weight $\omega$ in the training of the generator, we set it as 10.0.

We implement our model with Pytorch.We adopt the ResNet-50 \cite{he2016deep} pre-trained on ImageNet \cite{deng2009imagenet} as our CNN backbone. We modify the last layer stride to be 1 in the backbone Resnet50 to make final output features have more abundant information. Then we extract 2048d features for all images from the GAP layer.

\subsubsection{Reranking at test stage}
We use reranking \cite{zhong2017re} at the test stage to improve the model's performance. The key idea of reranking is, if a gallery image is similar to the probe in its k-reciprocal nearest neighbors, it is more likely to be a true match. We report results with and without reranking simultaneously. 

\subsubsection{Warming up learning rate}
We use a warming up strategy to adjust the learning rate at the train stage. It is a way to reduce the primacy effect in the early training stage. We adopt the linear warm-up strategy. The initial learning rate $R_{start}$ is used for the first $E_{warmup}$ epochs and will be linearly increasing from $\frac{R_{start}}{E_{warmup}}$ to $R_{start}$. We set the number of warm-up iterations $E_{warmup}$ as 10 in our experiments.

\subsection{Comparison with State-of-the-art Methods}

We exploit SYSU-MM01 to evaluate the accuracy of our model. We compare our results with current state-of-the-art RGB-IR cross-modality person ReID deep learning methods and the results are presented in Table \ref{tab:table1}. As shown in the table, our method (with or without reranking) is the best over all methods. It achieved at most 15.9 improvements on R1 and 14.4 on mAP with reranking (all search, single shot) and 7.4 improvements on R1 and 6.7 on mAP without reranking (all search, single shot).
This experiment demonstrates that our model is superior to state-of-the-art methods with respect to ReID accuracy.

\subsection{Ablation Study}

To illustrate the effectiveness of modules proposed in our work, we design six variant settings of our model. We define TS-GAN with ordinary discriminator as $\text{TS-GAN}_\text{{OD}}$ and TS-GAN with joint discriminator as $\text{TS-GAN}_{\text{JD}}$. The settings include the backbone module, $\text{TS-GAN}_\text{{OD}}$ with one cross-modality TS regularization loss, $\text{TS-GAN}_\text{{OD}}$ with all TS regularization loss, $\text{TS-GAN}_{\text{JD}}$ with all TS regularization loss and $\text{TS-GAN}_{\text{JD}}$ with the reranking scheme. The results of these settings in all-search single-shot are shown in Table \ref{tab:table2}.

\begin{table}[!h]
\centering
\caption{Results of different settings on SYSU-MM01.}
\begin{tabular}{lcccc}
\hline
\multirow{2}{*}{Methods} & \multicolumn{4}{c}{All-Search, Single-Shot} \\ \cline{2-5} 
 & R1 & R10 & R20 & mAP \\ \hline
$\text{Backbone}$ & 41.4 & 77.0 & 85.0 & 39.9 \\
$\text{TS-GAN}_{\text{OD}}^{\text{singleTS}}$ & 44.2 & 80.3 & 88.2 & 44.1 \\
$\text{TS-GAN}_{\text{OD}}^{\text{allTS}}$ & 47.6 & 82.6 & 89.9 & 44.5 \\ 
$\text{TS-GAN}_{\text{JD}}^{\text{allTS}}$ & 49.8 & 87.3 & 93.8 & 47.4 \\
$\text{TS-GAN}_{\text{JD,rerank}}^{\text{allTS}}$ & 58.3 & 87.8 & 94.1 & 55.1  \\ \hline
\end{tabular}%
\label{tab:table2}
\end{table}

From the table, we can see that the backbone module gets 41.4 R1 accuracy, which is directly trained on the SYSU-MM01 dataset with both RGB and IR modalities using triplet loss and cross-entropy classification loss.
The $\text{TS-GAN}_\text{{OD}}$ with single TS loss gets 44.5 R1 accuracy, 3.1 improvements to the baseline. This is because the cross-modality TS loss can reduce the domain gap between RGB and IR modality by making the middle layer feature maps of RGB input close to those of IR inputs. With the help of this scheme, the ReID backbone can learn IR information from Real RGB images. The $\text{TS-GAN}_\text{{OD}}$ with all TS loss obtains 47.6 R1 accuracy. This is because we add two more TS losses. Since the Teacher module can get very high accuracy, we make the middle layer feature maps of Real IR and Fake IR inputs through the ReID backbone module close to those of the same inputs through the IR Teacher module. These two loss functions can make the Student module learn similar features as the Teacher module so as to learn better representation. The $\text{TS-GAN}_\text{{JD}}$ with joint discriminator and all TS loss obtains 49.8 R1 accuracy, which is highest without reranking. The joint discriminator can further improve the image generation performance and person ReID accuracy by combining the ReID feature into adversarial training. Finally, all settings can improve R1 and mAP accuracy by using reranking.




\subsection{Analysis of Single Modality Results}

\begin{table}[!h]
\centering
\caption{Performance of pretrained model on single modality dataset.}
\begin{tabular}{cccccc}
\hline
Model     & Trainset  & Testset     & R1   & R10  & mAP  \\ \hline
Baseline & SYSU-IR  & SYSU-IR & 98.0 & 98.7 & 97.4\\
Baseline & SYSU-RGB & SYSU-RGB & 99.1 & 99.4 & 98.5 \\ \hline

\end{tabular}
\label{tab:table3}
\end{table}
Here, we evaluate the performance of the baseline on the single modality dataset. As shown in the table \ref{tab:table3}, we set the trainset and testset as the single modality part of SYSU-MM01.
Although the IR images have less information than RGB images, the baseline model can still work well on them. The reason is that when the trainset is under a single modality, there is no cross-modality feature gap. So we choose the IR model as the Teacher module because it is more natural to transfer RGB images to IR style.

\subsection{Visualization of Results}

\begin{figure}[h]
    \centering
    \includegraphics[width = 0.4\textwidth]{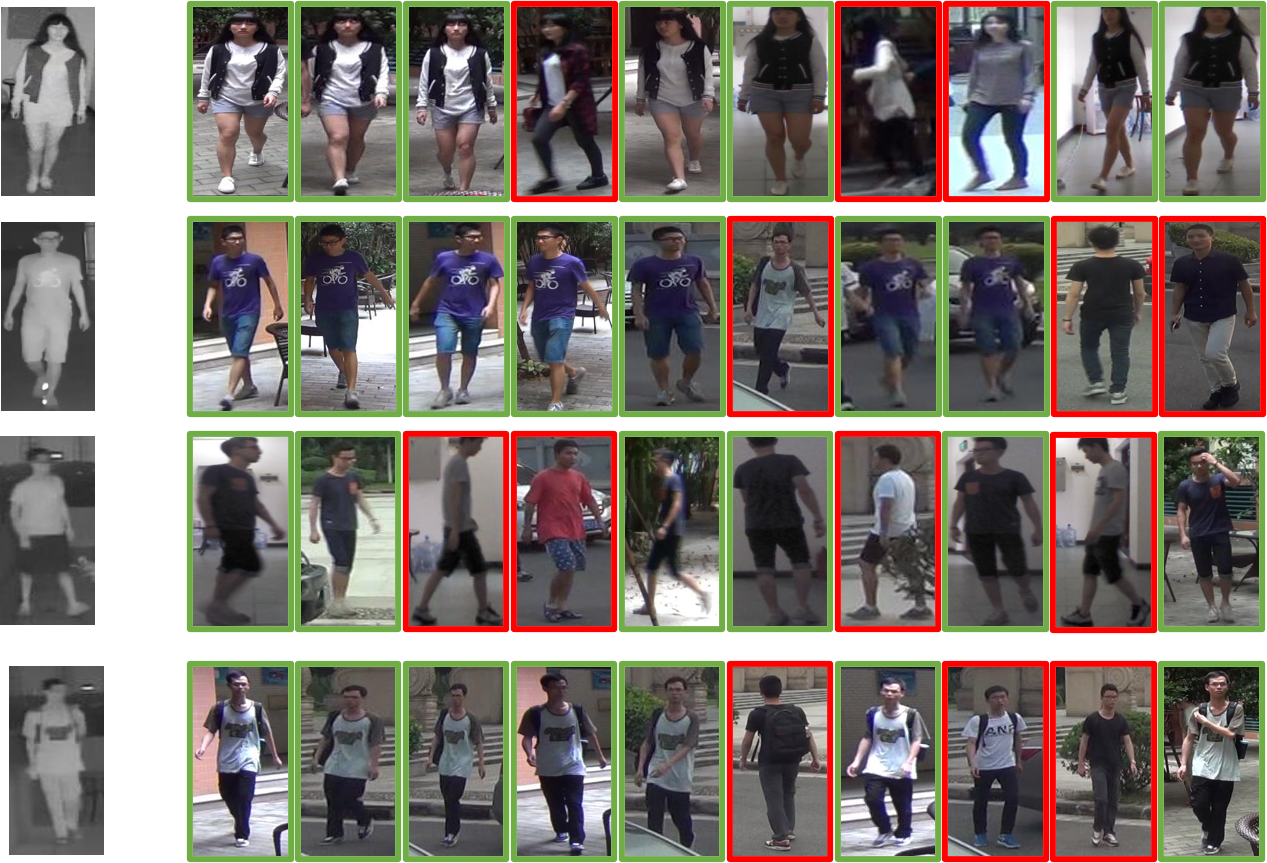}
    \caption{Examples of top-10 ranking results of IR query images. In each row, the images with green borders belong to the same identity as the query image and red ones are not.}
    \label{fig:Rank}
\end{figure}

We have visualized some results with four randomly selected query IR examples on the SYSU-MM01 dataset. The top ten retrieved results are shown in Fig \ref{fig:Rank}. The results showed that our proposed method is able to achieve good performance and minimize the domain gap between IR modality and RGB modality. However, there are still some situations not well handled. When the input IR person images lack large quantities of information, such as when the person faces away from IR cameras or the person image is highly blur, the model may fail to recognize the identity correctly.

\subsection{Visualization of Generated Images}

\begin{figure}[h]
    \centering
    \includegraphics[width = 0.4\textwidth]{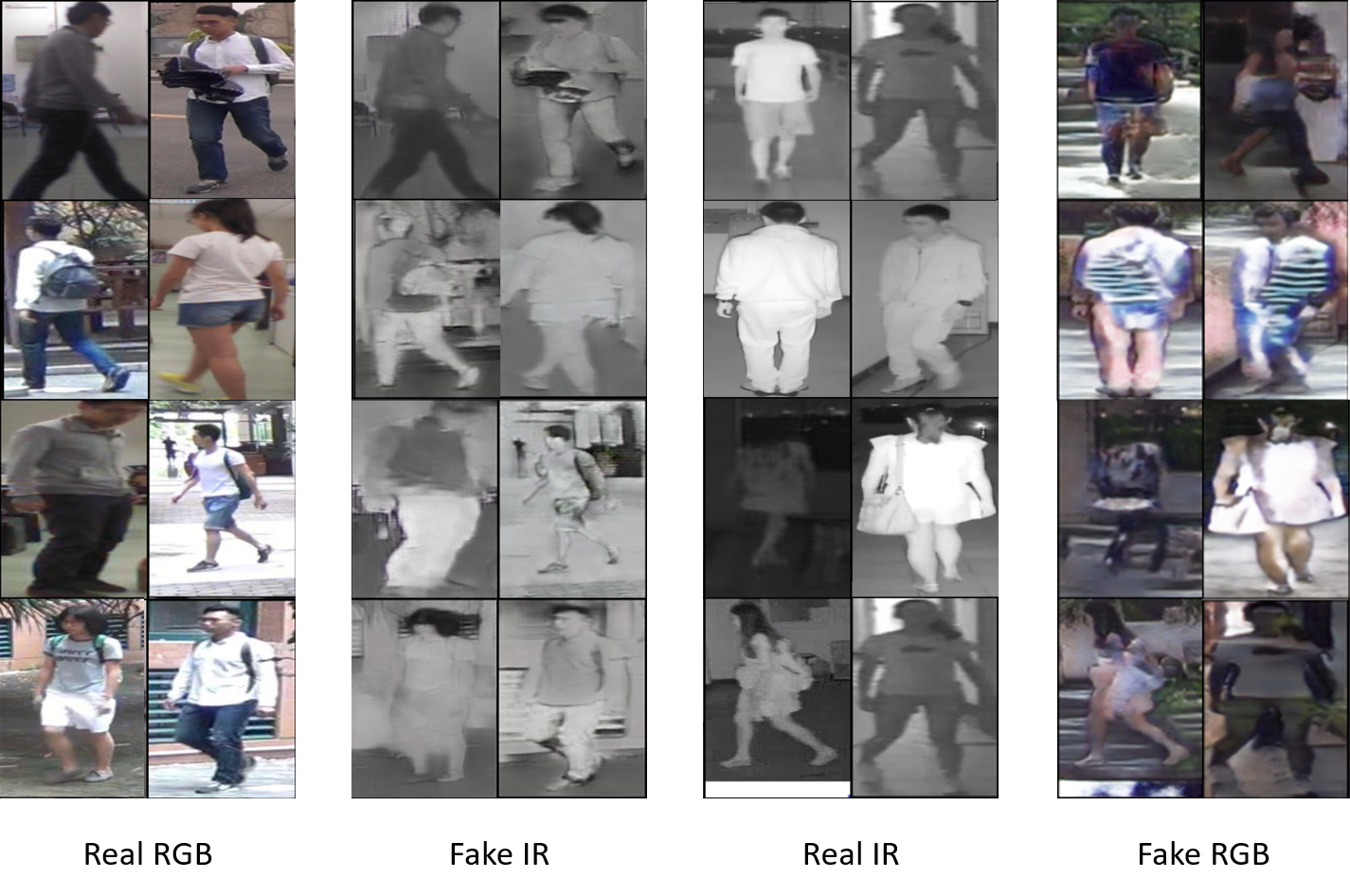}
    \caption{Examples of generated images on SYSU-MM01.}
    \label{fig:GEN}
\end{figure}

To show the performance of our generated images, we display examples of Fake IR images and Fake RGB images as well as their corresponding Real images. As shown in Fig \ref{fig:GEN}, Fake IR images look much clearer and more realistic than Fake RGB images. This is also the reason that we use the IR Teacher module rather than the RGB Teacher module.

\section{Conclusion}
In this paper, we proposed a novel TS-GAN model to learn the common representation features for RGB-IR cross-modality person images. In our method, TS-GAN consists of three main components, which are (1) IR generator and corresponding joint discriminator, (2) the ReID backbone and (3) the IR Teacher module. We adopted the ideas of GAN and TS to reduce the domain gap of inputs from different modalities in the ReID backbone. Comprehensive experiments on the challenging cross-modality person ReID dataset, SYSU-MM01, have demonstrated that our approach outperforms the state-of-the-art methods regarding ReID accuracy.
\bibliographystyle{IEEEtran}
\bibliography{refs}

\end{document}